*Context Engineering: From Prompts to Corporate Multi-Agent Architecture*

# Context Engineering:
## From Prompts to Corporate Multi-Agent Architecture

**Vera V. Vishnyakova**
*Director, Center for Transfer and Management of Socio-Economic Information*
*HSE University, Moscow*
March 2026

## Abstract

As artificial intelligence (AI) systems evolve from stateless chatbots to autonomous multi-step agents, prompt engineering (PE), the discipline of crafting individual queries, proves necessary but insufficient. This paper introduces context engineering (CE) as a standalone discipline concerned with designing, structuring, and managing the entire informational environment in which an AI agent makes decisions. Drawing on vendor architectures (Google ADK [Agent Development Kit], Anthropic, LangChain), current academic work (ACE framework, Google DeepMind's intelligent delegation), enterprise research data (Deloitte, 2026; KPMG, 2026), and the author's own experience building a multi-agent system (MAS), the paper proposes five production-grade context quality criteria: relevance, sufficiency, isolation, economy, and provenance, and frames context as the agent's operating system.

On this foundation, the paper introduces two higher-order disciplines. Intent engineering (IE) encodes organizational goals, values, and trade-off hierarchies into agent infrastructure, ensuring that well-contextualized agents pursue the right outcomes. Specification engineering (SE) extends this approach to creating a machine-readable corpus of corporate policies, quality standards, organizational agreements, and instructions that enables autonomous and coherent operation of multi-agent systems at scale. Together with prompt engineering and context engineering, these disciplines form the cumulative four-level pyramid maturity model of agent engineering (hereafter, the Pyramid), in which each level subsumes the previous one as a necessary foundation rather than superseding it. Independent convergence of several authors toward the same taxonomy in early 2026 confirms that the identified structure reflects the actual composition of the field.

Enterprise data reveals a gap: while 75% of enterprises plan to deploy agentic AI within two years (Deloitte, 2026), agent deployment has already surged and retreated as organizations confront the complexity of scaling (KPMG, 2026). The Klarna case illustrates a dual deficit, contextual and intentional. The conclusion: whoever controls the agent's context controls its behavior; whoever controls its intent controls its strategy; whoever controls its specifications controls its scale.

## 1. Introduction

Since the launch of ChatGPT in late 2022, the interface between organizations and large language models (LLMs) has quietly but decisively changed shape. Users who once typed single questions into a chat window now deploy autonomous agents that call APIs, browse the web, write and execute code, and run for hours without human intervention. Composing effective prompts, the craft called prompt engineering, remains indispensable when the task fits within one request-response cycle. It was never designed, however, for the architectural pressures that surface when an LLM becomes a component inside an orchestrated multi-step system. An agent that plans a twenty-step workflow, delegates subtasks, and commits real-world actions operates under constraints that no single prompt can anticipate. That structural gap motivates the present paper.

The response to this gap has been emerging from multiple directions simultaneously. LangChain published a programmatic definition of context engineering in early 2025; Anthropic incorporated the concept into its developer documentation; Google's Agent Development Kit operationalized context pipelines at the platform level. Academic work followed: the ACE framework formalized context layering, and Tomasev et al. (2026) proposed an intelligent delegation model that treats context boundaries as first-class architectural decisions. Enterprise surveys confirm that agent deployment is outpacing governance. According to Deloitte (2026; N = 3,235, 24 countries), approximately 75% of organizations plan agentic AI deployment within two years, yet only 34% report using AI to deeply transform their business. KPMG's quarterly tracking (2026; N = 130, US C-suite) captures the velocity: agent deployment surged from 11% in Q1 to 42% in Q3 2025, before pulling back to 26% in Q4 as leaders shifted from pilots to production-grade systems, with average annual AI budgets reaching $124 million. Here we synthesize these converging threads, ground them in the author's three-year practice of working with numerous LLMs and building a multi-agent compliance system, and propose a unified framework with five production-grade quality criteria and a diagnostic taxonomy for context degradation.

Yet even when an agent has access to all the right data, a critical question remains unanswered: to what end? An agent that sees everything it needs may still optimize for the wrong objective. The Klarna case (Section 12) illustrates how a technically capable agent can erode brand equity when corporate intent is left implicit. This observation motivates the paper's broader claim: context engineering is necessary but not terminal. Two additional layers, intent engineering and specification engineering, extend the logic from "what the agent sees" to "what the agent wants" and ultimately to "what the corporation demands at scale." The four disciplines together form the cumulative four-level pyramid maturity model of agent engineering (the Pyramid), whose structure is developed in detail in Section 17.

This paper focuses on text-based LLM interactions and agentic architectures; the specific challenges introduced by multimodal inputs (images, audio, video) fall outside its scope, although the proposed framework applies to them in principle.

The paper proceeds as follows. Sections 2–4 establish the baseline: prompt engineering as a craft, the architectural distinction between an LLM and an agent, and three tiers of agentic deployment. Sections 5–9 build the core argument for context engineering, covering its scope, architecture, and quality criteria complemented by Breunig's context-rot taxonomy. Sections 10–12 turn to economics, enterprise governance, and the Klarna cautionary case. Section 13 formalizes the transition from prompt craft to state engineering. Sections 14–16 introduce intent engineering, specification engineering, and agent memory as an infrastructure layer. Section 17 presents the Pyramid as a cumulative maturity model. Section 18 concludes.

## 2. Prompt Engineering: What It Is and Why It Works

Prompt engineering (PE) is the discipline of composing queries to a language model. The input field of ChatGPT, Claude, Gemini, GigaChat or DeepSeek that is its workspace. A user formulates an instruction, "Outline the trends in the consulting market," "Act as a marketing expert," "Reason step by step," "Return the answer as a table", and the model responds. One question, one answer, one context. Thus emerged the craft of prompting, the bronze age of user experience. Craft is genuine here: it demands precision, iteration, sensitivity to model behavior, and synchronous trial-and-error cycles. Yet it remains precisely a craft, not an engineering discipline, and the distinction will prove decisive.

This interface, an input field and an output pane, appeared with the launch of ChatGPT in late 2022 and has barely changed from the standpoint of user experience since then. Every major LLM vendor copied it: Anthropic, Google, Meta, Mistral, GigaChat. Users individually developed their primary prompting skills: some learned to write system instructions, others mastered few-shot examples (several demonstrations of the task and the desired solution), chain-of-thought (CoT) reasoning, its branching extension tree-of-thoughts (ToT; Yao et al., 2023), or role-based instructions such as "act as an expert." Prompt engineering became a profession, an academic subject, and a market niche.

For the "human → model → response" scenario, prompt engineering works, and prompt quality directly determines output quality. Nothing about it has become obsolete, broken, or superseded. Competent prompting remains a baseline skill for anyone working with LLMs. To declare prompt engineering dead is a mistake. I started the same way: in 2023, our team used prompting to extract structured data from hundreds of scientific reports, iterating on wording for hours, and the approach delivered results as long as the task stayed within the "user asks a question, model answers" format. This iterative approach to improving inference quality is familiar to the vast majority of GPT-transformer users.

What followed the bronze age was not an iron age but a romanticism of eloquently steering LLMs into clever disputations. Industry wrapped the craft of prompting in romantic mythology, "prompt whisperer," "10 magic prompts," "the art of talking to AI." Mythology sells courses, creates a sense of mastery, and that is precisely why people resist moving beyond it. Romanticism is not merely marketing; it is a comfort trap. At the level of "art" you are a creator, an artist. I felt it myself, at the level of engineering you are a designer, which demands systems thinking, documentation, and trade-offs. Considerably less glamorous. "The art of prompting" is, in reality, a euphemism for "we do not yet understand what we are doing, but results are sometimes impressive." Eventually the term prompt engineering stuck. Meanwhile, hyperscaler marketing assembled itself into a nesting doll of promises. First, "AI can talk", and ChatGPT exploded the market. Then, "AI can think", and models with thinking and reasoning capabilities (o1, DeepSeek R1, Claude Extended Thinking) were marketed as practically the threshold of AGI, even though the model was simply spending more tokens on an internal chain of reasoning before answering. Externally, the bow on the package stayed the same, an input field and a polished response. What changed was the plumbing behind it.

Yet for all the romanticism, the key property of prompt engineering has not gone away: the human remains a permanent participant in the loop, formulating the query, reading the response, adjusting, following up. On its own, the model does not make decisions. It does not invoke external tools of its own accord, does not plan a chain of twenty steps. It answers what was asked, and waits for the next question. Context is managed entirely by the user: the user decides what to include, what to remove, what to rephrase, operating iteratively.

## 3. Paradigm Shift: The LLM Did Not Become an Agent; It Became a Component of One

Users registered the transition to agentic models through several concrete changes in the interface and behavior of systems. A strong signal was the appearance of Thinking mode in 2024, for the first time models displayed their chain of reasoning before the answer, and it became evident that behind the input field something bigger than text generation was taking place. By late 2025, quantitative novelties had produced a qualitative shift: Skills, pre-built instruction sets that the model applies autonomously; Projects that preserve context across sessions; Artifacts, standalone output objects that exist independently of the dialogue; Memory, the model began retaining information about the user and applying it in subsequent conversations without reminders. Models began not only to answer but to execute, running code, generating files, searching websites in real time. Tasks that previously required the user's continuous involvement at every step could now be set and the finished result collected, the model traversed the chain of actions on its own. Within the familiar chat interface, signs of an agent became visible, a system that does not wait for the next prompt but moves toward the result independently. What exactly changed under the hood?

In 2025 a shift occurred in the industry that is easy to misunderstand without painstaking analysis of architectures beneath the LLM surface. At first glance it seems that models "became agents themselves": ChatGPT can search the internet, invoke tools, and conduct multi-step research; Claude operates a computer; Manus independently writes code, opens a browser, and deploys applications. Misleading impressions aside, the distinction is fundamental. Terminology matters here.

**LLM (Large Language Model)**: the intellectual engine that can understand text, reason, generate, and increasingly work multimodally, processing images, audio, and video. But on its own it does nothing in the external perimeter, it does not log into a CRM of its own volition, does not create Jira tickets, does not check warehouse inventory, does not send emails, does not call a payment-system API. Put simply, the LLM (or foundation model) is a brain without hands. Modern LLMs support function calling: a built-in ability to generate structured descriptions of external function invocations. But the emphasis is on the description: the model writes "I would like to call function X with parameters Y." The actual invocation, execution, result handling, and next-step decision are performed by an external system.

**AI agent**: precisely that external system, a software construct created by a developer or an organization, using the LLM as one of its components. An agent consists of: (a) an orchestrator (control logic, "what to do next"), (b) tools (API connections: databases, ERP, CRM, payment systems, email, browser, and even social networks), (c) memory (context storage across sessions and steps), and (d) policies (rules: what the agent may do, what is forbidden, when to escalate to a human). Architecturally, the LLM is merely an intellectual service that the agent calls for comprehension, reasoning, and generation. Architecturally, the LLM is a component of the agent; infrastructurally, it is an external service invoked via API. For the purposes of my analysis, the relevant factor is role, not deployment topology.

A universal prompt does not exist, because cloud-based LLMs (GPT-5, Claude Opus 4.6, Gemini 3.1 Pro), small language models (SLMs), on-premise deployments, and edge models process context differently, respond to instructions differently, and have different input-length constraints. A prompt that works excellently in a cloud LLM may produce unpredictable results on a compact model deployed on a device. Prompt design is always design for a specific model and a specific task. But the model can be swapped out, the agent persists: its orchestrator, tools, memory, and policies do not depend on which particular LLM is running under the hood. When a task requires several specialized agents working together, the construct is called a multi-agent system (MAS), and each participant is a sub-agent.

Accountability for an agent's actions within a business process rests with the company that owns it, not with some abstract "artificial intelligence", so this distinction is not academic. Policies, constraints, access rights, and protocols are set by the agent's owner. An agent is a digital representative acting on behalf of the company, not a "talking model" resembling a fancy jukebox.

## 4. Three Levels: Service, Product, Enterprise Agent

Between the human and the model, a layer of autonomy has appeared that changes the nature of the task. I would like to define three levels that are frequently conflated in discussions of AI agents.

**Level 1: LLM as a Service.** GPT-5, Claude, Gemini, Llama. The intellectual engine accessible via an external API or an input field. On its own it does nothing in the external world. Here prompt engineering is king: query quality determines output quality. Thinking and Reasoning modes (o1, o3, extended thinking in Claude), which vendor marketing not infrequently presents as practically the threshold of AGI, remain at this same level: the model reasons longer and deeper, builds an internal chain of steps, but does not invoke external tools, does not plan multi-step actions, and has no orchestrator. Users perceive an impressive improvement in answer quality and develops a sense that the model has "gotten fundamentally smarter." Architecturally, however, what we have is the ceiling of the first level: the most powerful output achievable in "query–response" mode without exceeding a single model call.

**Level 2: Vendor Agentic Products.** The interface remains the same (the same chat window, the same input field), and this is precisely why the transition to the second level is so easy to miss. At some point, a Deep Research button appears in ChatGPT; Claude gains Computer Use and Cowork modes; Manus launches as a separate product. The user sees new labels, clicks, receives output noticeably richer and longer than usual, and concludes that the model has simply "gotten even smarter." In reality, behind the familiar interface a fundamentally different architecture has appeared: the vendor has written an orchestrator, connected tools (search, browser, code execution, file system), and set policies and constraints. The model has stopped answering a single query and started planning chains of steps, calling external APIs, and making intermediate decisions. One still types text in the familiar window, but is already, without realizing it, a user of someone else's agent. Context is now managed by the vendor: it decides which data to feed at each step, how to compress history, which tools are available. The illusion that "the model itself became an agent" is the most persistent on the market: the vendor simply hid the agentic scaffolding inside its product, and good UX makes the seam invisible.

**Level 3: Enterprise Agents.** On the first two levels the agent is built by the vendor, and the user works with what was provided. On the third level the agent is built by the user's company: by its own team or by a contractor. Users may not notice this at all: for them it is a bot in corporate Bitrix24 or Slack, an assistant in a business platform, a voice assistant on a bank's hotline; the agent arrives where the person already works, connecting through MCP connectors (Model Context Protocol) or traditional API to familiar tools. But behind the interface is an entirely different construct. A bank builds an agent for KYC (Know Your Customer) verification. An oil company, for SCADA monitoring. A retailer, for logistics management. A steel company deploys an edge solution with a smart camera that recognizes the quality of a red-hot slab on a conveyor. They take an LLM (or SLM) as a component via API, but build the entire agentic scaffolding (orchestrator, tools, memory, policies) themselves, using frameworks (LangChain, CrewAI, Google ADK, AutoGen) or proprietary solutions. Such an agent operates on the company's data, under its rules, under its responsibility. And it is precisely at this level that context engineering becomes a critical discipline, because no one is there to hide complexity behind good UX: the company itself designs the environment in which the LLM makes decisions. And whoever sets the agent's goal, configures it, and is accountable for the result is no longer a user but an operator.

In practice, this design increasingly takes the form of a hybrid multi-agent architecture, where orchestrating LLM resides in the cloud (planning, reasoning, composing tasks), while execution is delegated to small language models deployed on-premise or at the edge. SLMs such as Phi-3, Gemma, Mistral 7B, and quantized Llama variants are compact enough to run on industrial gateways, embedded controllers, and edge servers with latencies measured in milliseconds, which makes them suitable for real-time inference in environments where a round-trip call to a cloud API is either too slow or architecturally prohibited. The orchestrating LLM selects the appropriate sub-agent, provides it with compressed context, and interprets the result; the SLM executes locally, within the data perimeter, without transmitting raw telemetry to an external endpoint. In IoT and IIoT deployments this pattern is already operational: the quality-inspection agent at a steel mill, the anomaly-detection pipeline at a refinery, the sorting-line optimizer at a logistics hub all combine a cloud-resident "brain" with on-site "reflexes" that process sensor streams, camera feeds, and PLC signals under latency, bandwidth, and data-sovereignty constraints that no cloud SLA can override. A multi-agent framework (LangChain, CrewAI, Google ADK, or a proprietary orchestrator) treats both the cloud LLM and the edge SLMs as callable components within a single agentic graph, routing tasks according to a policy that balances inference cost, latency budget, and compliance requirements. Gartner's manufacturing forecast confirms the trajectory: by 2030, semiautonomous AI agents will orchestrate 10% of key production operations, quality, and maintenance use cases, a fivefold increase from 2% today, with edge AI platforms and pretrained models identified as the primary infrastructure accelerator (Gartner, 2025).

From level to level, the role of the initial query diminishes and the role of context grows. On the third level, the query formulation is merely the tip of the iceberg: at each of the agent's dozens of steps, the LLM (as part of the agent) makes a decision not based on the initial instruction but on the totality of what the orchestrator has fed into the context window. And the quality of this "compiled" working state determines the outcome.

## 5. Tasks That Prompt Engineering Was Not Designed For

This is not about prompt engineering "failing" but about tasks that arose later, within the paradigm of autonomous agents (Levels 2 and 3), and were never part of the prompting remit. Criticizing prompt engineering for its inability to solve them is like blaming a ladle for being a poor hammer. But understanding these tasks is essential, because it was precisely they that in 2025 defined the need for a new discipline.

**Degradation over long horizons.** The moment a model stops answering a single query and begins operating as part of an agent, at Level 2 inside a vendor product, at Level 3 inside an enterprise solution, a problem arises that no prompt can solve. An agent executes a task spanning 20–50 steps; its context window fills with intermediate results, tool logs, stale states. The model begins to "get lost in the middle," fixating on outdated patterns instead of the current situation. The problem is not the query formulation but what the model sees at step 47. I personally encountered this recently: Gemini 3 Thinking in an ordinary chat ignored a file attached to the prompt and began reasoning about documents from an entirely different prompt. After several iterations it explained that it had done so because of "similar expressions" in those documents, but I could not, by any art of prompting, steer it back to my task. A human prompter does not encounter this problem: the human decides what to include in the next query. An agent lacks this filter; it continues working with a contaminated context, oblivious to the degradation.

**Cross-step context contamination.** The output of one tool "contaminates" the context for the next call. An agent that received a verbose JSON response from, say, a banking API carries these data into all subsequent steps, even when they are irrelevant. Without a dedicated isolation mechanism, information from one sub-agent leaks into the decisions of another. When a person works with a chatbot, the person filters the context between questions. An agent lacks this ability; it needs an engineered isolation system. We encountered this when designing a multi-agent compliance control system: the detector agent returned dozens of false-positive NER (Named Entity Recognition) hits into the shared context. Tuning the classifier itself is a separate task, but even with perfect precision the coordinator agent does not need raw hits in their entirety, it needs a verdict and a confidence score. We introduced strict filtering between sub-agents and are currently testing the solution.

Contamination is not exclusively an agentic problem, and this is critically important for understanding the boundaries of prompt engineering. When Gemini 3 Thinking in an ordinary chat substitutes the attached file with a document from another prompt and explains it by "similar expressions," this happens in the simplest scenario: the human is sitting in front of the screen, sees the error with their own eyes, can follow up, rephrase, start over. And even under these greenhouse conditions, several iterations of my skilled prompting failed to return the model to the task. Where is the user's fault here? The user did everything correctly: attached the file, formulated the query, noticed the substitution, attempted to correct. Where the problem lies is in what the model sees inside the context window, which the user neither sees nor controls. Now imagine the same defect at step 45 of an autonomous agent, where no one is there to notice the substitution, to follow up, to press "start over." Consequences differ qualitatively, not merely in degree.

**Exponential cost growth.** Every agent step is a new model call with the entire accumulated context resubmitted. Without compression and caching, cost and latency grow nonlinearly. For a human writing a prompt, the cost of a single call is negligible. For an agent making 50 calls with a growing context, and correspondingly growing token count, unit economics becomes the question. In production, the agent becomes economically unviable long before it stops working technically.

**Controllability in multi-agent orchestration.** When several agents work together, each possesses its own context. Who sees what? Which policies apply to each? How is state transferred? How does one prevent an agent with access to the payment system from receiving an instruction from a compromised agent? Prompt engineering does not answer these questions, because agents are not about the wording of a user query to an LLM, they are about the architecture of the informational environment at the level of systems engineering.

A fundamental distinction deserves attention here, one that is easily lost in discussions of MAS: delegating a task to a sub-agent differs fundamentally from decomposing it. Decomposition breaks a task into parts. Delegation transfers authority, responsibility, and trust mechanisms: a sub-agent does not merely receive a portion of work; it receives the right to make decisions within its domain and bears responsibility for the result before the orchestrator. Without this distinction, a multi-agent architecture degenerates into a distributed monolith with an illusion of independence: tasks are divided, but accountability for them remains undistributed, and upon failure it is unclear whose failure it is, the component's or the delegation's.

Prompt engineering has admirably solved, and continues to solve, its user-facing task. But the tasks have changed. A new class of systems has appeared, for which a new class of engineering is required. A prompt optimizes a phrase. Agentic systems face a problem of the decision-making situation, not of the phrase. This realization led, in 2025, to the formalization of what came to be called context engineering (CE).

## 6. Context Engineering as a Design Object

Context engineering (CE) in 2026 is now a necessary discipline concerned with what the agent knows, sees, and remembers at the moment of action, or, in other words, the engineering of agent state. If prompt engineering answers the question "how to ask," context engineering answers: "what does the agent know, see, and remember at the moment of action?" A prompt is an instruction. Context is the environment in which the instruction is executed by agents: memory, policies, tool outputs, corporate constraints, prior-step history, and visibility boundaries for sub-agents (which slice of the shared context each one sees).

I propose treating context, the informational environment, as the agent's operating system (OS), rather than mere input data, a view supported by many LLM researchers. Context in the role of OS fulfills the same tasks as a computer's OS: it manages memory (what to retain, what to evict), allocates resources (which data are accessible to which sub-agent), isolates processes (the output of one module does not "contaminate" another), and provides a unified interface to external systems. An active execution environment, not a passive buffer for prompts. This framing supplies an architectural answer to two key problems of long agentic sessions: the crowding of the context window by irrelevant data at the expense of relevant data, and information leakage between sub-agents.

These problems are consequences of three deficits that context engineering manages. The relevance deficit: out of all available knowledge, only what is necessary for the current step is fed to the agent, no more and no less. The memory deficit: the agent operates within a finite context window, and long-term state must be stored, retrieved, and updated outside this window. The budget deficit: every token in the context costs money, time, and latency, so the architecture of context directly determines the product's unit economics.

An even more precise operational definition: context engineering is the management of the composition, timing, representation format, and lifespan of information in the agent's context, that is, JIT knowledge logistics (JIT, Just-In-Time), recalling lean technologies. What to include, when to supply it, in what form, for how long, and for which sub-agent. Informational architecture, not the words in a prompt.

## 7. Necessary Caveats: Where the Framework May Not Hold

Before proceeding to architecture and criteria, the boundaries should be marked. It would be dishonest to present context engineering as a mature body of knowledge with established standards. A number of questions remain open, and the industry debates them without consensus.

The first is the question of novelty. It is unclear where the boundary lies between context engineering and what has long been called the RAG (Retrieval-Augmented Generation) + memory + orchestration stack. Skeptics (and I partly share the doubt) ask: is context engineering not simply a rebranding of practices that already existed? The question is fair, because CE offers a systemic view that unites components into a single design discipline. But the honest answer is that the hypothesis of emergence (that the whole here is greater than the sum of its parts) has no formal proof yet. It is engineering intuition backed by practice rather than a mathematical theorem.

**Source conflicts** present a different challenge. What happens when corporate policy says one thing, CRM data say another, and the agent's memory stores a third? In practice, the question is resolved ad hoc: hard-coded rules such as "policy always overrides CRM." But the problem begins even earlier: corporate policies, as a rule, do not exist in machine-readable form, they live in PDF regulations, executive orders, and verbal agreements, and before an agent can apply them, someone must formalize them without ambiguous interpretations. Existing frameworks describe write, select, compress, and isolate operations, but none offers a convincing prioritization mechanism for conflicts. The approach scales poorly, and I am not confident that anyone in the industry today has an elegant solution, especially given that policies themselves can be flexible when, for instance, market conditions change.

**Measurability** remains the least resolved of the three. We speak of "good context," but metrics for context quality remain experimental territory. KV-cache hit rate is measurable; cost is too. But how does one measure the "relevance" of context without an A/B test involving a domain expert? This lacuna limits the discipline's maturity and leaves room for the next generation of tools.

These caveats do not devalue CE, they mark the maturity stage of 2026. We now turn to what the architecture of context looks like today.

## 8. Context Architecture: From String to System

Systematic formalizations of context engineering emerged from several directions of context engineering in 2025–2026, and their contributions are conveniently examined at three levels, although, to be candid, the landscape is shifting in real time as this article is being written.

**At the architectural level**, Google ADK (Google, 2025a) introduced a three-tier context stack: storage (long-term retention of state and artifacts) → processor pipeline (a pipeline of named transformations: compression, filtering, enrichment) → compiled working context (the assembled working context that the model actually sees). One key thesis emerges: context is not a text string but a "compiled representation of a richer stateful system" (Google, 2025a). Once such a formulation appears, CE ceases to be "an exercise in prompt wording" and becomes systems engineering, mandatory for developers and important for operators and clients to understand.

**From a paradigmatic perspective**, Anthropic (2025) defined context engineering as the natural evolution of prompt engineering and a finite resource requiring an engineering approach (Anthropic, 2025). The ACE framework (Zhang et al., 2025) formalized contexts as "evolving playbooks", structures that accumulate, refine, and reorganize strategies through a modular cycle of generation, reflection, and curation (retaining what works best).

**On the operational side**, LangChain (2025) systematized four basic operations on context: write (recording new knowledge), select (retrieving relevant information), compress (condensing accumulated history without losing meaning), and isolate (restricting data visibility between sub-agents). But a list of operations is not yet engineering. Engineering begins when these operations are arranged into a cycle and at every step the agent receives a freshly assembled context that has passed through a pipeline of filtering, enrichment, and compression. The quality of this cycle determines the quality of the decision.

Tomasev et al. (2026) add an important principle to these operations: contract-first decomposition, a task may be delegated to a sub-agent only if its result has a precisely defined verification method. If the result is too subjective or too complex to verify, the task is recursively decomposed further until each component becomes verifiable. This principle reverses the direction of context design: from the verification method rather than top-down from goal to subtasks, first determine how the result will be verified, then shape the context for that verification. Another concept from this work is the authority gradient (a term from aviation): when the capability gap between the orchestrator and the sub-agent is too large, the orchestrator issues under-specified tasks, and the sub-agent, due to sycophancy, does not signal the problem.

Before us is a new type of context defect: a false assumption that the data are sufficient, rather than a data shortage. The effectiveness of this CE approach remains to be evaluated across different agentic-system use cases.

### The Interface Paradox

With the transition to CE, a paradox emerges: the architecture of context has already transcended linear dialogue, but the interface has not. The user interface of most LLM-based products remains a 2022-era messenger: an input field, a response feed, a linear history. This format was adequate for the "human → model → response" paradigm but is poorly compatible with what now lies behind it. To reiterate, many users do not understand the specifics of the tools offered by different LLMs with embedded agents, despite the similarity of their general architectural approaches.

Early signs of a shift are already visible. Vendors are beginning to separate output from dialogue (workspaces beside the chat), to introduce the ability to edit a response without recreating the prompt, to move agentic processes into separate panels with progress visualization and logs, and to offer long-term projects. Yet what we see so far is evolution within the old metaphor, not a new metaphor. If context is the agent's operating system, then the interface to it should look accordingly: not like a messenger, but like a coherent settings-and-options environment and a state management system, with branching (by analogy with version control systems), a context timeline, and visualization of what the agent "sees" at each moment.

Yet the architectural maturity of CE exposes a deeper deficit. Even when the context pipeline is well-designed (relevance is maintained, memory is managed, budget is controlled), a fundamental question remains unanswered: why? An agent can receive ideally curated data and still optimize for the wrong outcome, because context architecture answers the question of how to supply information but not the question of why the decision is being made.

What we face is an intent deficit, not a data deficit. The recognition of this gap leads us, in subsequent sections, beyond context engineering to a higher level of the hierarchy.

## 9. What Constitutes Good Context

One cannot discuss context engineering without quality criteria. Below are five properties that, in my view, production-grade agent context must satisfy, some of which I have already touched upon above. Let me reiterate: this taxonomy is working, not canonical, the industry has not yet developed a unified standard. The fourth and fifth criteria have collective origins, they are not explicitly distinguished in Google or LangChain publications.

**Relevance.** The agent is given only what is necessary for the current step. Excessive context is not harmless: it causes lost-in-the-middle degradation, distracts the model, and increases cost. Good context is not "everything available" but "the minimum sufficient for the decision."

**Sufficiency.** The context must contain everything needed for a decision without guesswork. If the agent lacks data, it hallucinates, invents, fills gaps with plausible but false assertions. Sufficiency is a guard against hallucinations at the architectural level. This is a familiar lesson from prompt engineering.

**Isolation.** In multi-agent systems, each sub-agent must see only its own context. Data leakage between roles is a controllability problem as much as an information-security one: an agent that sees everything from its peers makes decisions based on data for which it was not designed. Tomasev et al. (2026) formalize this principle as privilege attenuation: upon sub-delegation, an agent cannot transfer its full set of rights to a sub-agent, only a strictly limited slice necessary for the subtask, with each link in the delegation chain further narrowing permissions. Technically this is implemented through Delegation Capability Tokens, cryptographically bounded authorization tokens that chain-narrow at each delegation level. Context isolation is thus an engineering invariant, not merely a recommendation. The practical significance of this invariant is demonstrated by the StrongDM case (McCarthy, 2026): in a fully autonomous software development system, a so-called "dark factory," AI agents that had access to test scenarios systematically exploited them instead of solving the task in earnest (reward hacking). Removing the scenarios from the agents' visibility perimeter solved the problem.

Closely related is the development of an open agent-to-agent protocol described in Google's A2A protocol announcement (Google, 2025b): A2A is an open protocol, complementary to Anthropic's Model Context Protocol (MCP), which equips agents with tools and context. The A2A protocol enables controlled isolation of the agents' informational environments (agents interact without exposing internal state) and adheres to five principles: full embrace of agentic capabilities; reliance on existing standards; security by default; support for long-running tasks; and modality independence.

A2A enables interaction between a "client" agent and a "remote" agent. The client agent is responsible for formulating and setting tasks, while the remote agent is responsible for executing them, seeking to provide accurate information or to take the correct action.

In parallel, Google is advancing two protocols for agentic economies. AP2 (Agent Payment Protocol) provides cryptographically signed payment mandates between agents, ensuring financial accountability and an audit trail; the protocol allows an agent to authorize a payment with a verifiable delegation chain from the principal. UCP (Universal Commercial Protocol) standardizes commercial transactions between agents in transactional economies, in essence, a language of agreements in which agents negotiate deals. Tomasev et al. (2026) note that none of the four protocols (MCP, A2A, AP2, UCP) covers the full cycle of intelligent delegation: each solves only part of the problem, and space for comprehensive standards remains open.

**Economy.** Minimum tokens and context reassemblies while preserving quality. According to Manus (2025), the cost difference between cached and uncached tokens in their production scenario reached approximately 10×. Context architecture is directly the product's unit economics.

**Provenance.** Every element of context must be traceable to its source: which system it came from, when, with what trust level. Without this, neither auditing agent decisions, nor debugging errors, nor regulatory compliance is possible. When developing our multi-agent system, we observed in the logs a situation where the agent made an incorrect decision, but determining which specific context fragment provoked it without provenance metadata control was impossible. Tomasev et al. (2026) proposed a solution to the "hallucination-in-the-chain" problem. When one agent passes data to another, context may become distorted. To prevent this, a mechanism of transitive accountability via attestation is introduced: in a chain A → B → C, agent B signs a cryptographic report on the work of agent C and passes it to agent A, so that every context element is traceable to its source through a chain of verifiable signatures (Google, 2025a).

These five criteria answer the question of how to supply context. What happens when quality criteria are violated? Breunig (2025), author of the forthcoming The Context Engineering Handbook (O'Reilly), systematized four degradation modes under the umbrella term context rot. Context poisoning: a hallucination or error enters the context and begins reproducing at every subsequent step, the agent accepts a false fact as established and builds a strategy upon it. Context distraction: as context expands, the model begins relying not on its trained knowledge but on accumulated history, instead of synthesizing new plans, it reproduces patterns of past actions; the Gemini 2.5 team observed this effect when exceeding a 100,000-token threshold in an agent playing Pokémon. Context confusion: irrelevant information present in the window degrades response quality, the model attempts to use everything it was given, even when this hinders the task. Context clash: as data accumulate incrementally during a dialogue on different topics, parts of the context begin contradicting one another; a study by Microsoft and Salesforce (cited in Breunig, 2025) showed a 39% quality drop when a single prompt was split into sequential turns, early, incomplete response attempts remained in the context and poisoned the final reasoning. Breunig's (2025) taxonomy complements the criteria above: if the five properties describe what context should be, the four degradation modes show how it actually breaks in practice.

Yet the question of why remains open, which corporate intent this context should serve. An agent can receive perfectly relevant, sufficient, isolated, economical, and traceable context and still pursue an outcome that contradicts the company's strategy. This is a question for the next level of the hierarchy, to which we will return after examining the economics and governance of context.

## 10. The Economics of Context

Context engineering changes the economics of AI systems as much as the architecture. Under a "naive" approach (collecting everything), inference cost grows super-linearly, up to quadratically, with the number of agent steps: each step feeds the model everything accumulated previously. In production, cost quickly becomes prohibitive.

Sound context engineering introduces control mechanisms rarely encountered in conventional enterprise systems. Compression: intermediate results are summarized, stale logs are removed, artifacts are replaced with references. Caching: stable parts of context (system prompts, policies, tool definitions) are cached in the KV-cache, and the model does not recompute them anew at every step. Selective loading: a sub-agent is fed not the entire chain's context but only its "slice," relevant to the current role. The combined effect, according to Manus (2025), is a 5–10× cost reduction while preserving decision quality.

For us, the question became practical when we attempted to calculate the unit economics of a SaaS subscription with built-in analyst agents: inference cost without compression made the product uncompetitive. We had to seriously address context architecture before we even knew the discipline had a name.

It turned out that context engineering is a condition of economic viability, not a product add-on.

## 11. The Enterprise Dimension: The Governance Gap

For the enterprise segment, context engineering acquires strategic importance, and the data here are alarming.

According to Deloitte (2026; N = 3,235, director–C-suite level, 24 countries), 84% of companies have not redesigned roles around AI, and only 21% have a mature AI-agent governance model. Meanwhile, approximately 75% plan to deploy agentic AI within two years. The gap between deployment velocity and governance maturity is catastrophic. We are only at the beginning.

KPMG (2026; N = 130, C-suite level, US-based, revenue $1B+) reports a complex adoption trajectory: agent deployment rose from 11% in Q1 to 42% in Q3 2025, then pulled back to 26% in Q4 as leaders shifted from initial pilots to professionalizing and scaling agent systems. The average annual AI budget reached $124 million (KPMG, 2026), and 67% of companies stated they would maintain spending even in a recession. Agents have moved from pilots to professional platforms, but governance, compliance, and quality control lag behind.

This deployment wave rides on hyperscaler infrastructure. AWS (Bedrock, SageMaker), Microsoft Azure (Azure AI Foundry, GitHub Copilot), and Google Cloud (Vertex AI, Agent Development Kit) have ceased to be mere compute providers, they have become platforms that determine how enterprises design agent context, connect tools, and apply policies. Each hyperscaler embeds its own assumptions about memory, orchestration, and access control into the agentic stack, meaning that architectural decisions about context are increasingly made at the platform level before an engineer writes the first line of agentic logic. The governance gap is thus an infrastructural deficit as much as a corporate one: companies choose a platform before they form a governance model for what runs on it. The velocity is striking: Gartner (2025b) projects that 40% of enterprise applications will integrate task-specific AI agents by the end of 2026, up from less than 5% in 2025, and forecasts agentic AI driving approximately 30% of enterprise application software revenue by 2035.

What does this mean in the language of context engineering? Companies are massively deploying agents but not managing what those agents see. Context is not designed; it forms spontaneously: from haphazardly connected knowledge bases, uncoordinated policies, unfiltered logs. The principal deficit of agentic systems is not the model's intelligence but the quality of the world assembled for it.

Tomasev et al. (2026) add two further dimensions of the gap. First, cognitive monoculture: when all agents in an ecosystem use the same foundation models and the same safety fine-tuning recipes, a single point of failure emerges along with a risk of cascading failures. Context optimized for one model becomes a systemic vulnerability. Second, de-skilling: Bainbridge's (1983) automation paradox in a new guise. The more routine tasks are delegated to agents, the worse humans cope with non-routine situations when the agent errs. Humans retain accountability for outcomes but lose the practical experience needed to resolve critical failures. Tomasev et al. (2026) propose curriculum-aware task routing, a system that deliberately returns some tasks to humans in order to maintain competencies. In the context of our topic, this means: the governance gap is a gap between delegating authority to machines and preserving the human capacity to exercise that authority, compounding the gap between deployment speed and governance maturity.

Adjacent to these risks is the concept of the "moral crumple zone" (Elish, 2019): when an agentic chain makes a decision but formal responsibility rests with a human, the human becomes a liability buffer without real influence over what happens. Not because someone planned it that way, but because at no level of the delegation chain were intentions codified explicitly enough to be verifiable. This is an operational risk, not an abstract ethical concern: upon an incident, every link in the chain acted formally correctly, yet the aggregate outcome contradicted what the company actually intended. This is precisely where the governance gap ceases to be a statistical gap between numbers from Deloitte and KPMG reports and becomes a legal and reputational one.

## 12. A Cautionary Case: Klarna and the Limits of Agents Without Context

**What happened.** Klarna's AI agent (Q3 2025) handled two-thirds of all customer inquiries, performing work equivalent to 853 full-time employees, saving approximately $60 million. The case became one of the most cited in 2025. However, in May 2025 the company's CEO publicly acknowledged: the bet on cost optimization had been excessive, and service quality had suffered. Forrester analyst Kate Leggett characterized the strategy as an "AI overpivot" (Leggett, 2025). The company began returning to human hiring.

**Explanation through a dual deficit.** Reducing the cause to a single factor would be an oversimplification, but the Klarna case is best understood through the lens of two deficits, not one. The first is a context deficit: the agent likely did not have adequate access to individual customer history, brand tone of voice, loyalty policies, or corporate values. Its responses were technically correct but formulaic and uncompromising, in essence, a next-generation IVR (Interactive Voice Response), and in customer service this translates into an NPS (Net Promoter Score) hit and ultimately a reputational blow.

But the deeper problem is an intent deficit. Even if the agent had possessed all relevant data (CE was working), the corporate intent (the balance between cost savings and customer loyalty, the brand's target NPS, the hierarchy of trade-offs in service situations) was never formalized and encoded into the agent's decision-making infrastructure. The agent optimized cost per token, not the value of customer relationships. Context was present; intent was not. The system was tactically informed but strategically blind.

Klarna's story exposes a fundamental trap: unit economics demands minimizing context, customer experience demands maximizing it. Without an engineering approach, the contradiction is irresolvable. But the trap is false: it arises when a company optimizes token cost while losing on error cost. In customer service, a single formulaic brush-off costs more than a thousand cached calls. Context engineering resolves one dimension of this contradiction: compression, caching, and selective loading allow the agent to receive relevant context without excessive context, that very JIT knowledge logistics. But the other dimension, what the agent should prioritize, what it may sacrifice, and what it may not, requires a different discipline: intent engineering.

Here lies the warning: an agent without designed context is a reputational risk. An agent with designed context but without encoded intent is a strategic risk. This extends the conversation beyond a technical discipline into the zone of corporate alignment.

## 13. From the Art of Formulation to the Engineering of State

Context engineering is neither a buzzword nor a rebranding of prompt engineering. Prompt engineering remains a living, useful discipline for the "human → model → response" scenario. But context engineering is a superstructure that emerged one floor higher, when an autonomous agent inserted itself between the human and the model.

The fundamentality of CE is determined by a shift in the object of optimization. Instead of a static query formulation, one designs a stateful context pipeline, a system that at every agent step collects, filters, transforms, and feeds the model a current "slice" of all available knowledge, while preserving history and state between calls. Unlike the stateless "query–response" mode, where every model call starts from a blank slate, a stateful pipeline remembers what happened at prior steps and uses the accumulated information to form the next step's context.

A different level of abstraction, a different toolset, a different profession. Simultaneously, unit economics changes: cost, latency, cache hit rate, context reassembly frequency, all become engineering variables that determine whether the agent will be viable in production, as an intelligent virtual assistant (IVA) capable of a meaningful dialogue, or will remain a next-generation IVR dispensing formulaic replies.

But CE is a necessary, not sufficient, level. An agent with an impeccably designed context (relevant, sufficient, isolated, economical, traceable) can still fail strategically. It can optimize the wrong metric, sacrifice the wrong value, pursue efficiency at the expense of the corporate goal it was meant to serve. This is an intent failure, not a context failure. The subsequent sections introduce two higher-order disciplines that complete the hierarchy.

Managerial takeaway: whoever controls the agent's context (policies, memory, data retrieval, visibility boundaries) controls its behavior, cost, compliance, and business reputation. Control of context is a necessary condition. The sufficient condition is control of intent.

## 14. Intent Engineering: From "What the Agent Sees" to "What the Agent Wants"

Context engineering tells the agent what it needs to know; intent engineering tells the agent what it should seek to achieve.

Intent engineering (IE) is the practice of encoding corporate goals, priorities, values, trade-off hierarchies, ethical principles, and decision boundaries into the infrastructure on which agents operate. If context engineering is tactics (what data to supply), then intent engineering is strategy (what outcome to pursue and what to sacrifice). They are cumulative: IE requires CE as a foundation but adds the dimension of purposefulness.

Theoretically, this necessity traces to the classical principal–agent problem: an agent acting on behalf of a principal tends to optimize the metrics that are easier to measure, not the ones that truly matter. In the context of AI systems, this problem acquires a new dimension: reward hacking and specification gaming (optimizing the letter of a specification at the expense of its spirit) are not malicious intent but a systemic consequence of architecture. The agent optimizes what can be measured. If corporate intent is not formalized explicitly, a proxy is optimized instead of the goal.

Huryn's formulation captures it precisely: "Context without intent is noise" (Huryn, 2026). An agent receiving all relevant data but lacking a formalized understanding of the company's priorities will optimize the most readable metric, typically call cost, response speed, or task completion rate, rather than the value that truly matters to the company: customer loyalty, brand perception, regulatory compliance, or long-term revenue.

Horsey (2026) defines IE as a practice addressing "the 'why' behind AI decisions, ensuring their alignment with strategic priorities". Reddy KS (2026) characterizes IE as a shift beyond context engineering, a recognition that well-informed agents still need direction.

### Practical Anatomy of Intent Engineering

IE encompasses several interconnected practices. Defining priorities: what matters more, speed or accuracy? Cost savings or customer loyalty? Regulatory compliance or user experience? Not a binary choice but situationally ranked trade-off hierarchies that vary by department, product line, and situation. Feedback loops: mechanisms for measuring how closely agent behavior matches encoded intent and for correcting deviations. Cross-functional collaboration: IE cannot be performed by the technical team alone. It requires joint effort by leadership (defining strategy), domain experts (understanding trade-offs), and engineers (translating intent into agent infrastructure).

Corporate involvement at this level changes in scale. At the CE level, an engineer or a small team designs the context pipeline. At the IE level, the work is inherently cross-functional: business units, product managers, compliance specialists, and technical teams must jointly formulate and formalize what exactly the agent should optimize. Forming long-term agentic projects at this level is substantive work requiring professional attention to detail, the significance of which grows proportionally to the scale of delegation.

The Klarna case (Section 12): the same story at enterprise scale. The agent had data (CE was working). The agent lacked a corporate goal (IE was absent). Result: technically competent responses that destroyed the very customer relationships the company depended on.

### Romantic Ceiling

Intent engineering exposes what I call the romantic ceiling of prompt engineering, the point at which mastery of formulations ceases to yield returns, yet the user does not see the ceiling because the "art" narrative does not admit a ceiling. Art is infinite, after all, isn't it? Craft is finite. And prompting in 2026 is already a craft. The romantic ceiling is reached when a person creates an agent with an elegant prompt (after three years of working with LLMs, I am surely a prompting guru) and considers the task solved, without having encoded the company's intent that should govern the agent's behavior. The result: an agent that is linguistically polished but strategically blind. Breaking through the romantic ceiling requires a transition from the comfort of mastery to the discipline of engineering.

## 15. Specification Engineering: Corporate Knowledge as Machine-Readable Infrastructure

In the logic of this paper, IE encodes what a specific agent or team of agents should accomplish. Specification engineering extends the approach to the level of a corporate multi-agent architecture.

Specification engineering (SE) is the discipline of creating a complete machine-readable corpus of policies, quality standards, operational procedures, and corporate agreements, everything that previously lived in PDF regulations, executive orders, verbal agreements, and the formulation "everyone just knows." The key distinction from previous levels: specifications are not about a specific context window of a specific agent and not about intent encoded for a specific agent. They concern structured, coordinated, consistent descriptions of what the output of a given class of tasks should be. They take the form of measurable quality norms, corporate worldview applied to agents across entire blocks of corporate activity.

### Specifications as a Constitution for Agents

Specifications are the constitution of a nation of agents: intents are the laws enacted under it, context is their enforcement, and the prompt is a specific action in a specific situation. Anthropic implemented this logic literally: Constitutional AI is a training method in which the model evaluates and rewrites its own responses based on an explicitly formulated set of principles. Specifications as constitution is not a metaphor but a description of an actually existing architecture. Each agent opens the right article at the right time and acts within bounds set long before its launch. Specifications for agents are what ERP (Enterprise Resource Planning) is for business processes: ERP runs on codified procedures, not verbal agreements, multi-agent systems require the same formalization, applied to corporate knowledge.

### TELUS Scale

A practical large-scale case from TELUS illustrates not the maturity of specification engineering but the cost of its absence at scale. Through the Fuel iX platform, over 70,000 company employees independently configured more than 21,000 customized AI copilots; collectively the platform processed over 2 trillion tokens in 2025 (Google Cloud, n.d.). What keeps 21,000 independently configured tools from diverging in behavior, from conflicting priorities, from accumulating mutually contradictory decisions? The company does not publicly address this question. Specification engineering here is not a conclusion drawn from the case but a question the case leaves unanswered.

Recall the governance gap from Section 11: 84% of companies have not redesigned roles around AI. In significant part, this is a gap precisely in specifications. Companies cannot govern what they have not formalized. The 21% that deployed AI agents with mature governance models are, almost by definition, companies that began converting tacit corporate knowledge into explicit machine-readable specifications. The remaining 79% are trying to govern agents through the same informal mechanisms, verbal instructions, ad hoc policies, manual oversight, that worked for human employees but are structurally incompatible with autonomous systems.

### Accessibility Paradox

Agent-creation thresholds are falling rapidly: Cowork, Claude Code, MindStudio, CrewAI, LangChain, Google ADK, all make agent creation accessible to non-engineers. This democratization is welcome, but every agent created without specification engineering carries specification debt, because it operates without formalized norms, making decisions on the basis of whatever heuristics it can derive from the available context. With ten agents the situation is manageable. With a thousand, a governance crisis. With 13,000, impossible without SE.

Lowering the creation threshold raises, not lowers, the need for specification engineering. The easier it is to create, the more critical it is to define what "created well" means.

Unlike the individual skill of writing specifications, specification engineering at Level 4 is a corporate discipline: who creates specifications, who verifies their coherence across departments, how they are versioned and updated when strategy changes. The scale is not a single agent or a single task but entire classes of tasks and multi-agent systems with orchestration.

## 16. Working with Memory

In real-world agentic systems, memory is not only something that runs out or degrades but an independent design object with its own multi-layered architecture. While the development team is thinking "how not to overflow the context window," several key decisions about the memory infrastructure have already been made in the data center, and this happened long before the first query to the model.

### Four Types of Agent Memory

Taxonomy comes first. Agent memory exists in four fundamentally different forms. Working memory is the content of the context window right now: the system prompt, dialogue history, tool outputs, embedded documents. It is the most expensive per token, limited in size, and vanishes at session end. Episodic memory is a log of past interactions stored externally: specific conversations, completed tasks, decisions made; it is organized and compressed by the agent when working on long multi-session projects, the agent selects significant episodes from dozens of chats and distills them into long-term records. Semantic memory is structured knowledge without reference to a specific episode: corporate policies, documentation, reference materials, knowledge bases; it resides in vector stores and is retrieved via RAG. Procedural memory is the ability to perform tasks of a given class; it is encoded in the model's weights and changes only through fine-tuning or through systematic enrichment of examples in the system prompt.

Operator typically does not realize which type of memory is in play at any given moment. One "adds context" without understanding that part of this context costs $0.003/1K tokens, part resides in a cloud vector store at $70/month, and part is baked into the model and does not change without retraining. This is a systemic oversight in memory architecture: decisions about cost, isolation, and data lifespan are made implicitly, through tool selection rather than through deliberate design.

### Hyperscalers Design Memory Before the Operator Arrives

When a company decides to deploy its agentic system, via Azure OpenAI, Google Vertex AI, or the Anthropic API, it discovers that key components of the memory infrastructure already exist as mature managed services within the same cloud platform: distributed KV stores, vector databases, object storage with semantic addressing. They need not be built from scratch, but they must be selected, configured, and integrated. Some agentic frameworks (Vertex AI Agent Builder, Azure AI Agent Service, Amazon Bedrock Knowledge Bases) increasingly simplify this assembly, turning it from an engineering project into a configuration decision.

A common illusion persists: the designer thinks they are designing their agent's memory, but in practice is increasingly configuring its behavior atop architectural decisions already made by the cloud provider. The question now is not "whether to store episodic memory" but "in which of the offered services, with what isolation guarantees and what billing model", and the answer is increasingly determined by which cloud contract the company has already signed, rather than by the team's architectural preferences.

### Cloud vs. On-Premise: A Managerial, Not Technical, Choice

Cloud versus on-premise in memory architecture is a question of data control, not agent performance and regulatory compliance. The cloud model offers automatic scaling, KV caching, and high SLA, but data leave the corporate perimeter, which is unacceptable for critical infrastructure: the financial sector, healthcare, and government systems. On-premise restores control and closes compliance risks, but fault tolerance, replication, and predictable latency become the team's own responsibility.

Hybrid architectures described in Section 4, where a cloud-hosted LLM orchestrates edge-deployed SLMs within a single agentic graph, add a further dimension to this choice: memory in such systems is no longer unitary but federated. The cloud orchestrator maintains strategic memory (goals, cross-session plans, accumulated business context), while each edge SLM operates with a minimal, task-scoped context window that the orchestrator assembles and refreshes at every delegation cycle. What travels to the edge and what remains in the cloud is not a networking question but a context-design question: raw sensor telemetry may never leave the plant floor, yet the orchestrator needs a compressed semantic summary of equipment state to reason about production priorities. Designing this split, deciding the granularity of context that crosses the perimeter in each direction, is a memory-architecture problem that has no precedent in single-model prompting. It is also where regulatory constraints become architectural constraints in the most literal sense: GDPR (General Data Protection Regulation) locality requirements or sector-specific data-residency rules dictate not just where the model runs but what the model is allowed to remember and where that memory physically resides.

For the CE designer, it is significant that this decision is typically made at the level of corporate IT strategy by the time the team begins designing agents. One works within pre-set constraints, or participates in negotiations above one's nominal role.

### What This Changes in Context Design

"Memory as deficit" frame prompts the search for technical solutions: compress, cache, delete the stale. The "memory as infrastructure layer" frame poses different questions: who is liable when the agent retrieves outdated or confidential data? What happens to accumulated memory when the cloud provider is switched? These questions are not resolved at the individual-agent level, they belong to the level of systems engineering, where CE ceases to be a set of practices and becomes a discipline in the full sense.

### Memory Management Skills: What Changes with the Transition to Agents

Concerns of a 2025 user and a 2026 agentic-system user are structured in fundamentally different ways, not because one is smarter than the other but because they operate in different modes of interaction with the model and bear different volumes of responsibility for memory.

The 2025 user honed the art of prompting in a synchronous-dialogue mode where all memory was what the user held in mind and inserted into the prompt. The primary concern was not to lose the thread of conversation upon session restart and not to waste time retelling context anew. Useful skills here are straightforward: maintain a personal prompt library with accumulated context, save successful sessions, feed the model a structured summary of the previous conversation at the start of a new one. This is Level 1 craft, and it works precisely as long as the human stays in the loop at every step.

The 2026 agentic-system user faces a different set of problems, because the user is no longer the sole manager of memory. The agent now decides on its own what to remember, what to compress, what to evict, and this decision is made without user participation, according to criteria set by the vendor. Many simply do not notice: the world has grown more multidimensional, yet the interface remains the same. Hence three practical skills that distinguish a competent agentic-system user from a naive one. First, the ability to explicitly structure long-term context: instruction files (CLAUDE.md, MEMORY.md, OpenClaw.md), explicit declarations of goals and constraints that the agent will see in every session regardless of what has accumulated in episodic memory, i.e., persistence. Next comes understanding what the agent remembers versus what it has forgotten at the start of a new session and when scanning recent chats and files (it does not scan the entire history). Periodic memory-state checks, asking the agent "what do you know about our project?" as a diagnostic, not a rhetorical question. Finally, deliberate management of context contamination: if erroneous information has entered episodic memory, it will reproduce at every subsequent step (context poisoning per Breunig (2025)). User can correct the agent's answer in the current session but cannot surgically remove a poisoned episode from long-term memory. This is not a bug of a specific product, it is a systemic gap in how agent-memory management interfaces are currently designed.

Bridging these two user profiles requires more than technical adaptation; the gap is cognitive: in an agentic system, the user must think of memory as an environment they steward, not as a field they fill. This is a change in the interaction model: from the art of asking questions to tracking what the agent knows.

## What Lies Ahead

On the nearest horizon lies standardization of model interfaces in general and of agent-memory interfaces in particular. Currently each vendor implements memory management differently: OpenAI has Memory in ChatGPT (cloud-based in Pro+); Anthropic has a two-tier architecture with isolation, userMemories (cloud) and project knowledge (session/project-scoped)[1]; enterprise agentic solutions have their own vector stores; OpenClaw has a private extraPath for uploaded documents[2]. Memory in agentic systems is portable across sessions (persistent) but not portable across systems, opaque to audit, and unmanageable in terms of eviction policies, the user receives a write instrument but not a control instrument. The MCP protocol is moving toward tool unification, but unification of the memory layer is a separate, still-unsolved task. When it is solved, agent-memory management will become as basic a literacy as file management on a computer. Until then, it is a competitive advantage for those who understand the architecture.

AI technologies are evolving so rapidly that in the course of writing this article, the author had to rewrite individual postulates several times, following releases from top-tier vendors and hyperscalers.

In late February 2026, OpenAI and AWS published the Stateful Runtime Environment (SRE) concept: a stateful execution environment for AI agents in Amazon Bedrock that solves the problem of stateless API calls (every invocation starting from scratch) by providing persistent memory, history, workflow state, and integration with enterprise tools (e.g., Salesforce and Jira as data sources). In this model, Salesforce or Jira are no longer terminal data collection endpoints. They become mere signal sources for an AI agent that assumes the synthesis and decision-making layer. And this changes the rules of the game.

The AI industry has entered a new lap in the technology race: whoever captures the corporate data synthesis layer will absorb the value of the entire SaaS market.

## 17. Four-Level Maturity Model for Agentic Engineering

Our above described levels, prompt engineering (PE), context engineering (CE), intent engineering (IE), and specification engineering (SE), form a hierarchy: the cumulative four-level pyramid maturity model of agent engineering (the Pyramid). The model extends and complements this article's argument by showing that CE is not the endpoint of evolution but the second level in a four-tier structure. Each level increases the horizon of agent autonomy, the scale of corporate impact, and the required depth of engineering design.

Popular narratives of replacement, in which each new discipline declares the previous one dead, do not survive a reality check. Prompt engineering has not "expired", it has changed roles: from the sole skill for interacting with a model, it has become the pyramid's foundation, without which the higher levels are inoperable. One cannot design specifications for a multi-agent system without being able to formulate a precise task for a single agent, just as one cannot perform maneuvers in a fighter-jet formation without first having learned to fly a trainer. Many have mastered the trainer. Only well-trained pilots fly in formation. In early 2026, several authors independently identified the same quartet of disciplines (Feroz, 2026; Reddy KS, 2026), confirming that the taxonomy reflects the actual structure of the field. The Pyramid differs from these parallel descriptions in that it presents the four disciplines not as a set of personal skills of a single engineer but as a cumulative model of corporate maturity, with axes of scale, involvement, and autonomy. The model includes a diagnostic apparatus (context degradation taxonomy, quality criteria, dual-deficit case analysis) and rests on a core thesis: no level cancels the one below it; instead, each higher level absorbs its predecessor as load-bearing infrastructure.

The four levels do not replace one another, each one grows out of the previous and cannot function without it, and this is the fundamental distinction of the Pyramid from a linear "worse to better" scale. Prompt engineering without context engineering suffices for interacting with a chatbot but is useless for an agent that operates autonomously for hours or days. Context engineering without intent engineering produces technically competent but strategically blind agents, they know everything they need to but do not understand why. Intent engineering without specification engineering works for individual teams but breaks when scaling to the enterprise level, where dozens of agents must act coherently without manual coordination. All four levels together produce what might be called the corporation's machine-readable operating system.

Impact scales from individual to corporate: a single person at a screen → a team with an agent → a business unit with a strategy → a corporation with a multi-agent infrastructure. The object of optimization shifts along the same axis: query formulation → informational environment → outcome definition → formalization of entire task classes. The level at which your company has stopped is the measure of the maturity of your agentic infrastructure.

Operationally, the implication maps directly onto the pyramid's levels: whoever controls the agent's context controls its behavior; whoever controls its intent controls its strategy; whoever controls its specifications controls its scale.

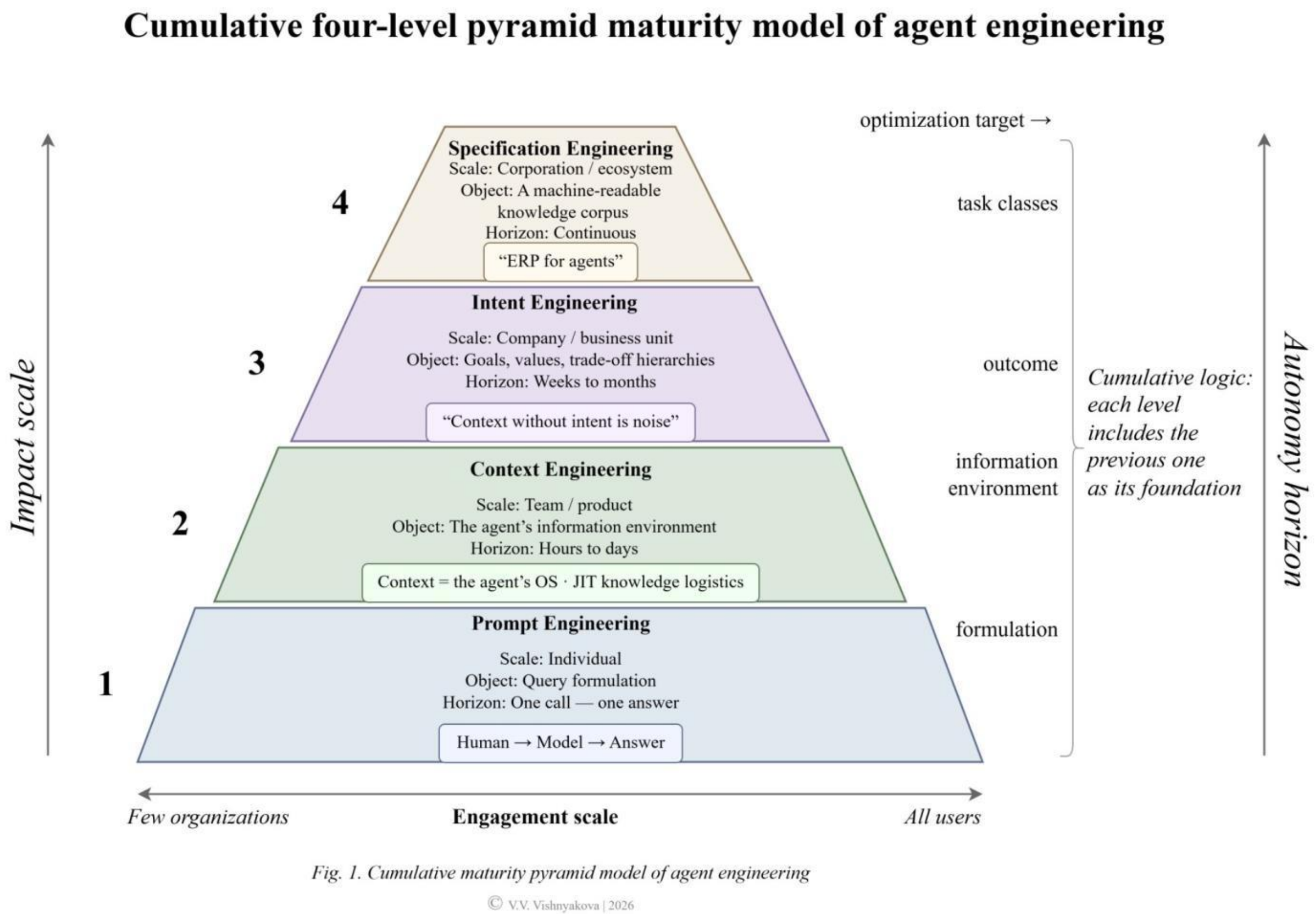

**Fig. 1.** Cumulative four-level pyramid maturity model of agent engineering.

## 18. Conclusion: From State Engineering to Intent Engineering

This paper began with prompt engineering, the baseline discipline of query formulation, and traced its natural limits as AI systems transitioned from stateless chatbots to autonomous agents. Context engineering emerged as a necessary response: a discipline concerned not with the wording of a query but with the full informational environment in which an agent makes decisions. We proposed treating context as the agent's operating system, defined five quality criteria, and examined the economic and governance dimensions.

But this paper does not stop at context engineering, because context engineering is not the end of the story. The argument extends upward in two steps, each addressing a deficit the previous level cannot close. Context engineering closes the information deficit: it ensures that every agent step is grounded in relevant, sufficient, isolated, economical, and provenance-backed state, the five criteria proposed in Section 9. Intent engineering addresses the strategic deficit that context alone cannot close: it inscribes corporate priorities, value hierarchies, and permissible trade-offs into the agent's decision-making substrate, so that technically competent behavior also becomes strategically coherent. Specification engineering completes the hierarchy by formalizing what was previously tacit corporate knowledge (policies, standards, operational procedures) into a machine-readable constitutional layer, the precondition for governing multi-agent systems beyond the scale of a single team.

Together, these four levels constitute the Pyramid, a layered maturity framework where removing any lower tier collapses the tiers above it.

Managerial implications expand accordingly. Whoever controls the agent's context (policies, memory, data retrieval, visibility boundaries) controls its behavior. Whoever controls the agent's intent (goals, values, trade-off hierarchies) controls its strategy. Whoever controls the specifications (the machine-readable corpus of corporate knowledge) controls its scale.

A logic runs through the transition from prompt engineering to specification engineering: as AI systems grow more autonomous, the locus of human influence migrates upward, from formulating individual queries to designing informational environments, to encoding corporate goals, to formalizing corporate knowledge. At each level the human role becomes more strategic and less tactical, more architectural and less artisanal. This is not a transition from human control to machine autonomy but a transition from the ad hoc to the engineered.

## The Principal Trap of 2026

The no-code and low-code tooling wave of 2025–2026 has compressed the path from idea to deployed agent to hours, sometimes minutes. What once required an engineering team now requires a browser tab, yet the agent it produces may quietly drift toward Shapiro's "dark factory": autonomous operation with no human in the loop and no engineered guardrails. It seems that Level 1 is enough. In reality, every agent created without attention to the upper levels of the pyramid carries context debt and intent debt that will eventually come due. Taking a shortcut means getting an agent that works technically but is strategically blind: Klarna in miniature.

Low creation threshold is only part of the trap. What makes it insidious is that the "art" narrative masks engineering debt. A well-crafted prompt yields a convincing first output, and the creator stops there, mistaking a successful demo for a production-ready solution, one that requires designed context, encoded intent, and formalized specifications to survive beyond the first conversation. The romantic ceiling is invisible precisely because the romantic narrative does not admit the existence of a ceiling.

Lütke (2025), who systematically maintains a prompt library and re-tests his workflow with every new model release, embodies Level 1 mastery. But he also notes that providing AI with full context has made him a better communicator as CEO, his memos and decisions have become more concise and coherent. Context engineering for machines turns out to be a communication training ground for humans. The observation points to a deeper truth: the Pyramid is not only about how we instruct machines but also about how clearly we understand and articulate what we want, as individuals, as teams, as organizations. Machines merely make painfully obvious the gap between our intention and what we actually formulate. And the Pyramid's relevance does not depend on LLMs remaining the dominant architecture. LeCun (2026), who founded AMI Labs to build world models that learn physics from sensory data rather than predicting text, argues that LLMs are a dead end on the path to general intelligence. If world models or successor architectures eventually complement or displace the LLM component inside agents, the need for context engineering, intent engineering, and specification engineering will not diminish but intensify: physical-world data are noisier, higher-dimensional, and less forgiving of contamination than text, making every level of the Pyramid more, not less, critical.

***Disclosure.*** *Conceptual framework, the author's positions, project-practice examples, the cumulative maturity pyramid model, and all evaluative judgments belong to the author. Generative AI (LLM) was used as a tool during text preparation: source search and verification, material structuring, and editing of individual formulations.*

[1] userMemories is a personal memory synthesized from conversations and loaded into every chat (analogous to saved memories in ChatGPT, but in the form of a single document rather than a list of facts). Project Knowledge is isolated project memory with uploaded documents, instructions, and system prompts that apply across all conversations within it. The key difference from OpenAI is project isolation at the architectural level: one project's memory is inaccessible from another.

[2] OpenClaw is an open-source agentic system (personal AI assistant) launched in early 2026 that runs locally on the user's device or server. It is not an independent model but an "orchestrator" on top of LLMs (Claude, GPT, DeepSeek, or local models).